\documentclass{article}

\usepackage[accepted]{icml_2021}
\usepackage{natbib}

\usepackage[utf8]{inputenc} 
\usepackage[T1]{fontenc}    
\usepackage{hyperref}       
\usepackage{url}            
\usepackage{booktabs}       
\usepackage{amsfonts}       
\usepackage{nicefrac}       
\usepackage{microtype}      
\usepackage{graphicx}
\usepackage{subcaption}
\usepackage{amsmath,xspace}
\usepackage{multirow}
\usepackage{paralist}

\newcommand{\mouter}{\textsf{Outer}\xspace}
\newcommand{\inner}{\textsf{Inner}\xspace}
\newcommand{\latent}{\textsf{Outer+Latent}\xspace}
\newcommand{\baseline}{\textsf{Baseline}\xspace}
\newcommand{\outereq}{\textsf{Outer+EQ}\xspace}
\begin{document}

\twocolumn[

\icmltitle{An Integer Linear Programming Framework for Mining Constraints from Data}

\icmlsetsymbol{equal}{*}

\begin{icmlauthorlist}
\icmlauthor{Tao Meng}{ucla}
\icmlauthor{Kai-Wei Chang}{ucla}

\end{icmlauthorlist}

\icmlaffiliation{ucla}{Department of Computer Science, University of California, Los Angeles, USA}

\icmlcorrespondingauthor{Tao Meng}{tmeng@cs.ucla.edu}
\icmlcorrespondingauthor{Kai-Wei Chang}{kwchang@cs.ucla.edu}

\vskip 0.3in
]
\printAffiliationsAndNotice{}

\begin{abstract}
    Structured output prediction problems (e.g., sequential tagging, hierarchical multi-class classification) often involve constraints over the output label space. These constraints interact with the learned models to filter infeasible solutions and facilitate in building an accountable system. However, although constraints are useful, they are often based on hand-crafted rules. This raises a question -- \emph{can we mine constraints and rules from data based on a learning algorithm?} 
    
    In this paper, we present a general framework for mining constraints from data. In particular, we consider the inference in structured output prediction as an integer linear programming (ILP) problem. Then, given the coefficients of the objective function and the corresponding solution, we mine the underlying constraints by estimating the outer and inner polytopes of the feasible set. We verify the proposed constraint mining algorithm in various synthetic and real-world applications and demonstrate that the proposed approach successfully identifies the feasible set at scale. 
    In particular, we show that our approach can learn to solve 9x9 Sudoku puzzles and minimal spanning tree problems from examples without providing the underlying rules. Our algorithm can also integrate with a neural network model to learn the hierarchical label structure of a multi-label classification task. Besides, we provide a theoretical analysis about the tightness of the polytopes and the reliability of the mined constraints.
\end{abstract}

\section{Introduction}
\label{sec:intro}
    A variety of machine learning problems involve making coherent decisions over a set of output variables, where the dependencies between them can be described by constraints~\citep{punyakanok2005learning, samdani2012efficient, nowozin2011structured}. For example, in part-of-speech tagging, a constraint specifying that every sentence should contain at least one verb and one noun can greatly improve the performance \citep{ganchev2010posterior}. Similarly, in hierarchical multi-label classification, a figure labeled `flower' should also be labeled `plant' as well~\citep{dimitrovski2011hierarchical}. To incorporate constraints with learned models, one popular method is to formulate the inference problem into an integer linear programming (ILP)~\cite{roth2004linear}.  This framework is general and can cope with constraints formed as propositional logics \citep{hooker1988generalized, richardson2006markov}. 
    This approach has been widely used in natural language processing, computer vision, and many application areas. It has demonstrated great performance gains in various applications (e.g., \citet{martins2010turbo, nowozin2011structured,roth2004linear,goldwasser2012predicting,chang2008constraints,meng2019target}).\footnote{Solving ILP is in general NP-hard. However, in practice, inference problems often can be solved efficiently using a commercial ILP solver or an approximation inference technique (e.g., LP-relaxation~\cite{fromer2009lp}, loopy belief proprogation~\cite{murphy2013loopy}). See discussion in \cite{finley2008training}. }
    
    In the literature, existing works most focus on how to utilize constraints to facilitate learning. They 
    mostly assume constraints are given as a priori. 
    However, for some applications, manually identifying constraints is tedious. Besides, some constraints are obscure and cannot be easily identified by human experts.\footnote{For example, if we shuffle columns of all sudoku puzzles with the same order, the puzzles still follow a set of constraints. However, it is hard for humans to recognize these underlying rules.} Inspired by representation learning methods automate feature extraction, we envision that \emph{an artificial intelligence system that could automatically recognize underlying constraints among output labels from data and incorporate them in the prediction time.}
    
    To illustrate the goal, consider the following learning problem. We are given a set of input-output pairs as training data, where each input is an adjacency matrix of a graph representing the distances between nodes and the output is corresponding minimal spanning tree (MST). Our goal is to train a model to generate MST of a given graph (adjacency matrix) \emph{without} telling the model that the output is a tree. 
    Specifically, the model has to identify the underlying constraints that are satisfied by all training samples from a family of candidate constraints. The success in this problem has a great potential; however, there are limited prior works except some methods extending the basic Valiant's algorithm \citep{valiant1984theory}, such as inductive logic programming \citep{muggleton1994inductive, riedel2006incremental} and constraint learning \citep{bessiere2013constraint, bessiere2016new, bessiere2017inductive}. Most of them use logic clauses to formulate the constraints and solve the satisfiability problem. However, it is unclear how to incorporate them with machine learning models. 
    
    Inspired by the great success of ILP in constrained output structure predictions, we propose a novel framework to formulate the constraint learning based on ILP. 
    In particular, we estimate the feasible set defined by the constraints and 
    explore three techniques: 1) mining inequality constraints to form a superset of the feasible set by constructing an outer polytope based on seen data; 2) mining equality constraints by dimension reduction of the superset; and 3) mining complex constraints with a latent variable method. We also propose an algorithm to induce the subset of the feasible set for evaluating the quality of the constraints. Note that although the constraint mining algorithm is designed under the ILP framework, our algorithm does not involve solving ILP when mining the constraints.
    
    
    We evaluate the proposed framework on three tasks: MST, Sudoku, and hierarchical multi-label classification. The first two tasks demonstrate that our method is able to mine complex structures and deal with large label space. For example, in Sudoku, our model can perfectly learn the underlying rules and achieve $100\%$ accuracy. We then incorporate the proposed approach with a learned neural network on hierarchical multi-label classification.
    Our framework helps models learn the structure in label space and improve the performance by over $10\%$ compared with the baseline.
    Finally, we conduct a comprehensive analysis on MST. We verify the constraints learned by our approaches by comparing the corresponding feasible set with the ground truth. We also provide a theoretical estimation on the feasible set size and compare it with the empirical results and discuss the running time of the algorithm. The source code and data are available at \url{https://github.com/uclanlp/ILPLearning}.
    
\section{Related Work}
\label{sec:related}
    \paragraph{Constraints Formulated by Integer Linear Programming}
        ILP is widely used in formulating constrained inference in machine learning tasks, including
        semantic role labeling~\citep{vasin2004semantic},
        entity-relation extraction~\citep{roth2005integer}, sentence compression~\citep{clarke2008global}, dependency parsing~\citep{martins2009concise}, multi-lingual transfer~\citep{meng2019target}, corefernece resolution~\citep{chang2013tractable}, relation extraction~\citep{ye2020integrating} and reducing bias amplification~\citep{jieyu2017men}.
        These works use pre-defined constraints to formulate ILPs. In contrast, we aim to mine constraints from data.

    \paragraph{Mining Constraints From Data}
    \citet{raedt2018learning} summarize the milestones in constraint mining. Learning logical rules from data can be traced back to the Valiant's algorithm \citep{valiant1984theory} that mines the hard constraints formulated as $k-$CNF. Inductive logic programming \citep{muggleton1994inductive, riedel2006incremental}, as an extension of Valiant's Algorithm, is aiming to deal with general first-order logic. It has been used in both real world~\citep{bratko1994applications} and mathematical applications \citep{colton2006mathematical}. Constraint learning \citep{bessiere2013constraint, bessiere2016new, bessiere2017inductive} combines these two approaches  together. Besides, several efforts have been put on relaxing logical constraints such as soft constraint learning \citep{rossi2004acquiring}. \citet{wang2019satnet} use semidefinite programming (SDP) \citep{wang2019low} to relax the maxSAT problem and cooperate with deep learning (see Sec. \ref{sec:exp} for comparison). Another way to relax the logical constraints is to relax the Boolean variables to be continuous variables \citep{li2019logic, li2019augmenting}, or continuous random variables like probabilistic soft logic \citep{kimmig2012short, bach2015hinge, embar2018scalable}.
        Most of these previous works use logics to represent the constraints. In contrast, we design a framework based on ILP and use the linear form to formulate the constraints. This allows us directly incorporate constraints with inference in structured output predictions. Some concurrent works \citep{pan2020learning,tan2020learning} learn constraints by initializing a set of constraints and updating them based on gradient, but it is not able to obtain the guarantee that the constraints are tight and converge to the ground truth as ours do (see discussion in Sec. \ref{sec:method} and Sec. \ref{sec:analysis}).
        
\section{Mining Constraints with Integer Linear Programming}
\label{sec:method}
    We first review the constraint mining framework based on ILP. We then propose an approach to mine constraints by estimating the outer and inner polytopes of the feasible set. Finally, we discuss how to extend the framework to capture complex constraints.
    
        ILP is a linear optimization problem with linear constraints and the values of variables are restricted to integers. Formally, the ILP problem can be formulated as
        \begin{equation}
            \label{eq:ilp0}
                \max\nolimits_{\mathbf{y}\in\mathbb{Z}^d} \quad \mathbf{w}\cdot \mathbf{y} \qquad \mbox{s.t.} \quad A\mathbf{y}\leq \mathbf{b},
        \end{equation}
         where $\mathbf{w}\in\mathbb{R}^d$ is the coefficients of the objective function (a.k.a. weights) and $\mathbf{y}$ is an integer vector that encodes the output label.\footnote{In structure output prediction, usually each element of $\mathbf{y}$ takes value 1 or 0, indicating if a specific value is assign to a specific output variable or not. $\mathbf{w}$ are the scores of sub-components of output assigned by a model.}
        The matrix $A$ and vector $\mathbf{b}$ specify the constraints.  We use $S^*$ to denote the feasible set defined by the constraints. 
        Various structure prediction problems can be casted into the ILP formulation. For example, dependency parsing can be formulated as finding the maximum spanning tree in a directed graph~\citep{mcdonald2005non}, where each node represents a word and the edge $w_{ij}$ represents how likely the word $i$ is the dependent of the word $j$ predicted by a model.
       $\mathbf{y}=\{\mathbf{y}_{ij}\}, \mathbf{y}_{ij}\in \{0,1\}$ is the indicator of the edges in the resulting tree. 
       The objective in Eq. \eqref{eq:ilp0} then can be interpreted as the total score of edges in $\mathbf{y}$, and the constraints, described by $(A, \mathbf{b})$, restrict $\mathbf{y}$ to be a tree~\citep{martins2009concise}. 
      
        Prior works (see, e.g., \cite{martins2009concise}) mostly assume the constraints $(A, \mathbf{b})$ are given. 
        However, in this paper, we assume $(A, \mathbf{b})$ are unknown and our goal is to identify the underlying feasible set $S^*$ spanned by $(A, \mathbf{b})$ using a set of objective-solution pairs $\{(\mathbf{w}^{(i)},\mathbf{y}^{(i)})\}_{i=1}^k$ that satisfy constraints
        defined by $(A, \mathbf{b})$. 
For example, in MST, giving a set of weights $\mathbf{w}^{(i)}$ (adjacency matrix) with the corresponding optimal solution $\mathbf{y}^{(i)}$, our algorithm identifies the structure of the output $\mathbf{y}$ form a tree structure. 



        In the following, we introduce algorithms to estimate the feasible set for mining the underlying constraints. These mined constraints define an superset of the feasible set. We also design an algorithm to get the subset of the feasible set to evaluate the estimation. 
        
    \subsection{Mining Inequality Constraints}
    \label{sec:inequality}
        \begin{figure}[t]
            \centering
            \begin{subfigure}{0.47\linewidth}
                \includegraphics[width=\linewidth]{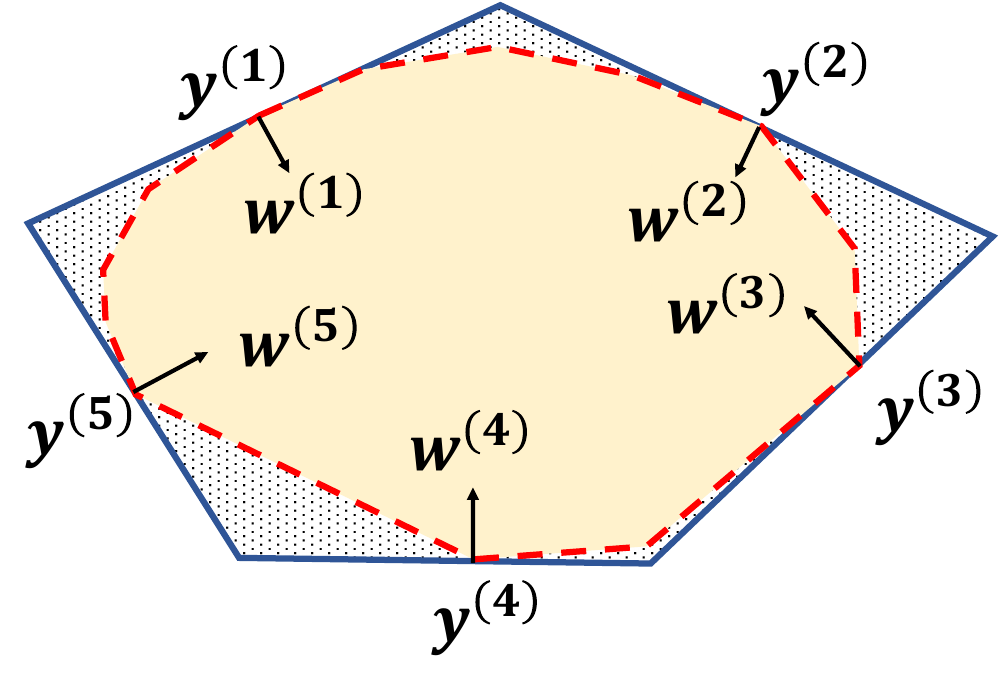}
                \caption{Outer polytope ($S_O$)}
                \label{fig:1a}
            \end{subfigure}
            \begin{subfigure}{0.47\linewidth}
                \includegraphics[width=\linewidth]{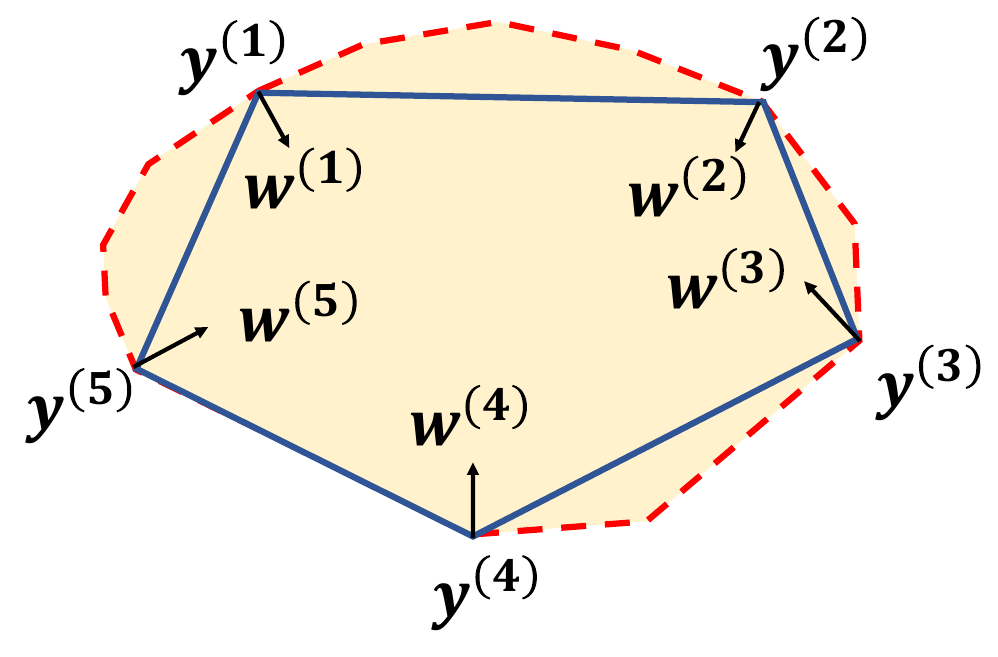}
                \caption{Inner polytope ($S_I$)}
                \label{fig:1b}
            \end{subfigure}
            \caption{The pentagons (blue solid line) in Fig. \ref{fig:1a} and \ref{fig:1b} show the outer and the inner polytopes of $5$ training samples $\{\mathbf{w}_i, \mathbf{y}_i\}_{i=1}^5$ (see Sec. \ref{sec:inequality}). The dashed red line shows the boundary of the feasible set (yellow region). We also show $w_i$ as the normal vector of the outer line. 
            }
            \label{fig:uandl}
        \end{figure}
    
        In the following, we discuss how to estimate the underlying feasible set $S^*$ associated with inequality constraints. Our approach finds a convex hull $S_O$ defined by a set of learned inequality constraints that is an outer polytope (i.e., superset) of the feasible set $S^*$. We also propose a method to get an inner polytope $S_I$ that is a subset of $S^*$ and use the gap between the $S_O$ and $S_I$ to estimate the quality of approximation. Figure \ref{fig:uandl} shows an example about $S_I,S_O$ defined by 5 training samples in a 2-dimensional space. 
        We denote $S_I^{(i)},S_O^{(i)}$ as the inner and outer polytopes after considering the first $i$ samples. 
        
        \paragraph{Outer polytope} We first introduce how to identify $S_O$. Assume that we already know part of the constraints $A', \mathbf{b}'$. We initialize the outer polytope as 
        $S_O^{(0)}=\{\mathbf{y}\in \mathbb{Z}^d\ |\ A'\mathbf{y}\leq \mathbf{b}'\}$ (if $A',\mathbf{b}'$ are empty, $S_O^{(0)}=\mathbb{Z}^d$).
        For every training sample $(\mathbf{w}^{(i)}, \mathbf{y}^{(i)})$, we consider adding the following constraint to the outer polytope
        \begin{equation}
            \label{eq:upperconstr}
            \mathbf{w}^{(i)}\cdot\mathbf{y}\leq \mathbf{w}^{(i)}\cdot\mathbf{y}^{(i)}.
        \end{equation}
        Since $\mathbf{y}^{(i)}$ is the optimal solution under weight $\mathbf{w}^{(i)}$, all the points in the feasible set must sit in the half-space defined by Eq. \eqref{eq:upperconstr}, otherwise $\mathbf{y}^{(i)}$ is not the optimal solution. We have
        $S_O^{(i)}=\{\mathbf{y}\in S_O^{(0)}\ |\ \mathbf{w}^{(j)}\cdot\mathbf{y}\leq \mathbf{w}^{(j)}\cdot\mathbf{y}^{(j)},\ j=1,2,\dots,i\}, \mbox{ and}$
        $S^*\subseteq S_O=S_O^{(k)}\subseteq \dots \subseteq S_O^{(1)} \subseteq S_O^{(0)}.$
        
        The outer polytope $S_O^{(i)}$ (the upper bound of the feasible set $S^*$) is tight when we only observe the first $i$ samples. That is, assuming $S_O^{(i)}$ is the feasible set, if we query $\mathbf{w}^{(1)},\dots,\mathbf{w}^{(i)},$ we will find $\mathbf{y}^{(1)},\dots,\mathbf{y}^{(i)}$ are (one of) the optimal solutions.  Therefore, $S_O^{(i)}$ is a possible feasible set. Since $S^*\subseteq S_O^{(i)},$ this bound is tight. This shows that without any further assumption, we cannot do better than $S_O$ for estimating the outer polytope of the feasible set $S^*.$
        
        In the test time, we are requested to conduct inference with unseen input weight $\mathbf{w}^{(q)}$. Since all constraints in $S_O$ are linear, we solve
        the following ILP problem 
        \begin{equation}
            \label{eq:ilp}
            \begin{split}
                \max\nolimits_{\mathbf{y}\in\mathbb{Z}^d} \quad &  \mathbf{w}^{(q)}\cdot \mathbf{y} \\
                \mbox{s.t.} \quad &
                \begin{bmatrix}
                    A'^T & \!\! \mathbf{w}^{(1)} & 
                    \!\!\dots\!\! & 
                    \mathbf{w}^{(k)}
                \end{bmatrix}^T
                \mathbf{y} 
                \\
                &\leq
                \begin{bmatrix}
                    \mathbf{b}'^T \!\!& 
                    \mathbf{w}^{(1)} \cdot \mathbf{y}^{(1)} &
                    \!\!\dots\!\! &
                    \mathbf{w}^{(k)} \cdot \mathbf{y}^{(k)}
            \end{bmatrix}^T\!\!.
            \end{split}
        \end{equation}
        
        The objective value of the solution of Eq. \eqref{eq:ilp} might be higher than the optimum as the solution might not satisfy all the underlying constraints. We will show that empirically the outer polytope can approximate the feasible set effectively in Sec. \ref{sec:exp}. Although the number of constraints grows linearly with the number of training samples, we find that empirically the inference time does not grow much.\footnote{In structure output prediction, constraints are often associated with only the problem structure. Therefore, all the inference instances share the same constraint set, and the overhead in solving ILPs is amortized~\citep{srikumar2012amortizing,kundu2013margin,chang2015structural}.}
        
        \paragraph{Inner polytope} To understand the quality of $S_O$, we also construct the inner polytope $S_I$, then we can use the gap between $S_O$ and $S_I$ to estimate the quality of the approximation.  
        We first initialize $S_I^{(0)}=\emptyset$. For every training sample $i$: $(\mathbf{w}^{(i)}, \mathbf{y}^{(i)})$, we set
        $S_I^{(i)}=convex\_hull(\{\mathbf{y}^{(1)},\mathbf{y}^{(2)},\dots,\mathbf{y}^{(i)}\}),$
        and then $S_I^{(i-1)}\subseteq S_I^{(i)}.$ Since all $\{\mathbf{y}^{(i)}\}$ are in the feasible set that is convex, all the convex hulls must be subsets of the feasible set. Therefore,  we have
        $S_I^{(0)}\subseteq S_I^{(1)} \subseteq \dots \subseteq S_I^{(k)}=S_I\subseteq S^*.$
        
        Similarly, we can prove that $S_I^{(i)}$ is a possible feasible set after observing the first $i$ samples, which means as a lower bound, $S_I^{(i)}$ is also tight.
        
        When we conduct inference with $S_I$, we examine every vertex of the convex hull and choose the one with the optimal objective. Since it is an inner polytope of the feasible set, the solution is guaranteed to satisfy all constraints, and the objective value can be lower than the optimum. Although inner polytope and outer polytope methods are two separate algorithms, the gap between their objective function value and the size of the feasible set provide an estimation of the tightness of the bound. 
        
        In Sec. \ref{sec:analysis} we will show that, empirically,  this approach converges with reasonable number of training samples and running time is discussed in Sec. \ref{sec:time}.
        
        \paragraph{Dealing with predicted weights} When we incorporate the proposed approach with a structured prediction model, the weights $\mathbf{w}$ are predicted by a base model.  
        In this situation, the predicted weights $\mathbf{w}$ can be noisy and the corresponding label $\mathbf{y}$ may not be the optimal solution to Eq. \eqref{eq:ilp}.
        As the result, the outer polytope may not contain some feasible solutions as they are filtered out later by the algorithm.  
        To handle the noise, we adapt Eq. \eqref{eq:upperconstr} to 
        \begin{equation}
            \label{eq:adaptconstr}
            \mathbf{w}^{(i)}\cdot \mathbf{y}\leq\mathbf{w}^{(i)}\cdot\mathbf{y}^{(i)}+\xi_i,\ i\in[k],
        \end{equation}
        where $\xi_i$ is a slack variable to ensure every training point $\mathbf{y}_i\in S^*$ satisfies Eq. \eqref{eq:adaptconstr}
        $$\xi_i=\min\nolimits_{j\in[k]}\{\mathbf{w}^{(i)}\cdot \mathbf{y}^{(j)}-\mathbf{w}^{(i)}\cdot\mathbf{y}^{(i)}\}.$$
        
    \subsection{Mining Equality Constraints}
    \label{sec:reductD}
        When there are equality constraints in the output label space. Effectively, the dimension of the output space is reduced. However, the dimension of the set $S_O$ is the same as that of $\mathbf{w}$ and $\mathbf{y}$. Therefore, this inspires us to find the sub-space of $S_O$ to further tighten the feasible set.
        
        For example, in the MST problem the number of edges we select is exact $N-1$ where $N$ is the number of nodes. Formally, the linear constraint $\mathbf{1}\cdot\mathbf{y}=N-1$ holds for every feasible point $\mathbf{y}.$ 
        
        We denote this $d'-$dimensional affine sub-space as $S_D=\{\mathbf{y}\ |\ W_{eq}\cdot\mathbf{y}=\mathbf{c}\}$. We can obtain $W_{eq},\mathbf{c}$ by solving the kernel of $[\mathbf{y}^T,\mathbf{1}],$ which is
        \begin{equation}
            \label{eq:kernel}
            \begin{bmatrix}
                \mathbf{y}_1^T & \mathbf{y}_2^T & \dots & \mathbf{y}_n^T \\
                 1 & 1 & \dots & 1 \\
            \end{bmatrix}^T
            \begin{bmatrix}
                W_{eq}^T \\
                -\mathbf{c}^T
            \end{bmatrix}
            =\mathbf{0}
            .
        \end{equation}
        The intersection of $S_O$ and $S_D$ is used to replace $S_O$ as the outer polytope of the feasible set: 
        $S_I\subseteq S^*\subseteq S_O\cap S_D=\{\mathbf{y}\in S_O\ |\ W_{eq}\mathbf{y}=\mathbf{c}\}.$
        For the reliability of this algorithm, we give two lemmas:
        \paragraph{Lemma 1:} Given training data $\{(\mathbf{w}^{(i)},\mathbf{y}^{(i)})\}_{i=1}^M$ that satisfy the ILP constraints set defined in Eq.\eqref{eq:ilp0}, if the ILP contains an equality constraint $W_{eq}\cdot\mathbf{y}=c$, our equality constraint mining algorithm can identify it.
        \paragraph{Proof Sketch:} For any underlying equality constraint $W_{eq}\cdot\mathbf{y}=c,$ all the labels of training points $\mathbf{y}_i$ should satisfy it and it must be in the kernel of Eq. \eqref{eq:kernel}.
        \paragraph{Lemma 2:} For an equality constraint given by this algorithm, the probability that this constraint does not hold for the optimal solution of a random query is less than $\frac{1}{eM}$, where $M$ is the number of training points.
        \paragraph{Proof Sketch:} We use $D_w$ to denote the domain of the query (weights), and $f^*(\mathbf{w})$ as the ground truth solution for the query $\mathbf{w}$. We denote the equality constraint learned by the algorithm is $g(\mathbf{y})=0.$ Therefore, all the data $\mathbf{y_i}$ in the training data satisfy $g(\mathbf{y}_i)=0.$ We let $p=Pr_{\mathbf{w}\sim D_w}(g(f^*(\mathbf{w})=0)).$ The probability of all the training data satisfying $g(\mathbf{y}_i)=0$ is $p^M$. Thus, the probability that this constraint does not hold for a random query solution is
        $$p^M(1-p)\leq \frac{M^M}{(M+1)^{M+1}}<\frac{1}{eM}.$$

    \subsection{Latent Variables}
    \label{sec:latent}
        Some prediction problems involve constraints with complex logics and require auxiliary variables to model the problem structure. Thanks to the flexibility of the ILP framework, we can introduce latent variables to extend the expressiveness of the constraint mining framework:
        \begin{equation}
            \label{eq:latent}
                \max_{\mathbf{y}\in\mathbb{Z}^d} \quad \mathbf{w}\cdot \mathbf{y} 
        \qquad \mbox{s.t.} \quad A_{pre}
                \begin{bmatrix}
                    \mathbf{y} \\
                    \mathbf{h}
                \end{bmatrix}
                \leq \mathbf{b}_{pre},
               \ \quad A
                \begin{bmatrix}
                    \mathbf{y} \\
                    \mathbf{h}
                \end{bmatrix}
                \leq \mathbf{b},\\
        \end{equation}
        where $\mathbf{h}$ are latent variables, and they appear in the constraints but not in the objective function in Eq. \eqref{eq:latent}. 
        Despite that $\mathbf{h}$ is not part of the output, it facilitates to formulate the ILP problem. In general, a set of pre-defined constraints $(A_{pre}, \mathbf{b}_{pre})$ are given to describe the relations between $\mathbf{y}$ and $\mathbf{h}$. Then, given a set of $\{(\mathbf{w}^{(i)},\mathbf{y}^{(i)})\}_{i=1}^k$, our goal is to learn the constraints $(A, \mathbf{b})$.
        
        The latent variables can help us formulate the constraints better in the ILP framework. Specifically, with the help of the latent variables, some inequality constraints can be reformulated as the equality ones. As it is easier to identify equality constraints in our framework, this will make the constraints we learn more accurate. We adapt the method in Sec \ref{sec:reductD} to solve the kernel of the matrix 
        $[\mathbf{y}^T, \mathbf{h}^T, \mathbf{1}]^T.$ 
        In this way, we can mine equality constraints with respect to $\mathbf{h}$ and then derive the outer polytope $S_O.$ Since $\mathbf{h}$ is determined by the variables $\mathbf{y}$, adding constraints on $\mathbf{h}$ also reduces the size of $S_O.$
        
        For example, consider multi-label classification with output $\mathbf{y}$, where $y_i, i = 1\ldots m$ is a binary indicator of class $i$.
        If we would like to identify the constraints between pairs of labels from $\{(\mathbf{w}^{(i)},\mathbf{y}^{(i)})\}$, we can introduce a set of latent variables $\{h_{i,j,b_1,b_2}\}_{i,j=1\ldots m; b_1,b_2\in \{0,1\}}\}$ with pre-defined equality constraints 
        $h_{i,j,b_1,b_2} = (y_i=b_1)\wedge (y_j=b_2), \forall i,j,b_1,b_2$.
        They can be further formulated as
        \begin{equation*}
            \forall i,j,\ 
            \begin{cases}
                h_{i,j,1,0}+h_{i,j,1,1}=y_i\\
                h_{i,j,0,1}+h_{i,j,1,1}=y_j\\
                \sum_{b_1,b_2 \in \{0,1\}}h_{i,j,b_1,b_2}=1.\\
            \end{cases}
        \end{equation*}
        
        By introducing $h_{i,j,b_1,b_2}$, we are able to capture some correlations between labels better. For example, label $i$ and label $j$ cannot be positive at the same time can be represented by an equality constraint $h_{i,j,1,1}=0$. Without latent variables, it can only be represented by inequality constraint $y_i+y_j\leq 1$.  In this case, the introducing of latent variables make the learned constraints more accurate.

\section{Experiments}
\label{sec:exp}
    We experiment on two synthetic problems, $9\times 9$ Sudoku and minimal spanning tree (MST), to show that the proposed methods can capture different kinds of constraints. We then incorporate the proposed technique with a feed-forward neural network in a hierarchical multi-label classification problem. For all the experiments, we use the Gurobi v8.1.1~\citep{gurobi} as the ILP solver.\footnote{We configure the ILP solver such that it outputs optimal solution (i.e.,
    set $MIPGap=0$). The relaxed LP will be discussed in Sec. \ref{sec:time}.}
    
    \subsection{Sudoku}
     \label{sec:sudoku}           
        \begin{table}[t]
            \caption{Sudoku results. EQ stands for equality constraint mining. Performance is reported in entry-level  accuracy. Our approaches can successfully identify the underlying Sudoku constraints. 
            }
            \begin{subtable}{\linewidth}
                \centering
                \caption{Original Sudoku}
                \begin{tabular}{c|c|c}
                    Model & Train & Test\\
                    \hline
                    ConvNet \citep{kyubyong2018sudoku} & 72.6\% & 0.04\% \\
                    ConvNetMask \citep{kyubyong2018sudoku} & 91.4\% & 15.1\%\\
                    SATNet \citep{wang2019satnet} & 99.8\% & 98.3\%\\
                    Outer + EQ (ours) & 100\% & 100\%\\
                \end{tabular}
                \label{tab:1-1}
            \end{subtable}
            \begin{subtable}{\linewidth}
                \centering
                \caption{Permuted Sudoku}
                \begin{tabular}{c|c|c}
                    Model & Train & Test\\
                    \hline
                    ConvNet \citep{kyubyong2018sudoku} & 0\% & 0\% \\
                    ConvNetMask \citep{kyubyong2018sudoku} & 0.01\% & 0\%\\
                    SATNet \citep{wang2019satnet} & 99.7\% & 98.3\%\\
                    Outer + EQ (ours) & 100\% & 100\%\\
                \end{tabular}
                \label{tab:1-2}
            \end{subtable}
            
            \label{tab:sudoku}
        \end{table}
        
        In Sudoku, given a $9\times 9$ grid with numbers partially filled in, the player is requested to fill in the remaining of the grid with constraints that each $3\times 3$ sub-grid, each column and each row must contain all the numbers of $1,2,\ldots,9$. The size of the feasible set is $6.67\times 10^{21}$ \citep{felgenhauer2005enumerating} which is extremely large. 
        Our goal is to use the proposed method to solve Sudoku puzzles without telling the model the rules. 
        
        We follow the experiment setting in \citet{wang2019satnet} to represent the solution of Sudoku as a vector $\mathbf{y} \in \{0,1\}^{729}$, where $y_{ijk}$ denotes the $i-$th row $j-$th column is the number $k$ or not. 
        The partially filled entries $(r_i,c_i)=n_i$ (i.e., row $r_i$ column $c_i$ is number $n_i$) are encoded in $\mathbf{w} \in \{0,1\}^{729}$, where we set the corresponding weight for $w_{r_ic_in_i}$ to be 1 and the rest to be $0$. In this way, maximizing the objective function $\mathbf{w}\cdot\mathbf{y}$  guarantees $y_{ijk}=1$ if $w_{ijk}=1$.
        Then given pairs of $\{\mathbf{w}, \mathbf{y}\}$, our methods mine the underlying constraints $A$ and $\mathbf{b}$ in Eq. \eqref{eq:ilp}.

        We experiment on the dataset introduced in \citet{wang2019satnet}. The dataset contains $9,000$ training and $1,000$ test samples, each of which has a unique solution. We also conduct experiments in the permuted setting \citep{wang2019satnet}, where a pre-defined permutation function is used to shuffle the $9\times 9$ grid. In the permuted setting, it is almost impossible for humans to identify the underlying rules, despite the puzzle is still filled in a certain order. We follow the configuration in \citet{wang2019satnet} to compare our approach with a convolution neural network for Sudoku~\citep{kyubyong2018sudoku} (\textsf{ConvNet}) and \textsf{SATNet}~\citep{wang2019satnet} and report our results along with their published results in Table \ref{tab:sudoku}.
        
     As shown in the table, by using the equality constraints mining technique, our framework can realize the Sudoku rules and achieve $100\%$ accuracy. The constraints we mine reduce the size of candidate solution space from $2^{729}$ to $6.67\times 10^{21}$ (i.e., the number of feasible Sudoku puzzles). Note that our approaches, as well as \textsf{SATNet}, do not utilize the position clues in the data. Therefore, it is not affected by permutations. 
       
       In fact, the equality constraints we learn are exactly the rules of Sudoku. The Sudoku rules can be represented linearly as
       \begin{equation}
        \label{eq:sudokuconst}
            \begin{split}
                \sum_{i=1}^9 y_{kij} = 1,\ 
                \sum_{i=1}^9 y_{jki} = 1,\ \forall j,k,\\ 
                \sum_{i=1}^3 \sum_{j=1}^3 y_{(x+i)(y+j)k} = 1, \forall x,y\in \{0,3,6\} \forall k.
            \end{split}
        \end{equation}
        We use $S^*$ to denote the space defined by Eq. \eqref{eq:sudokuconst} and $\hat{S}$ as the space of our constraints. We verify the $S^*=\hat{S}$ by
        \begin{compactenum}
            \item Verifying that each constraint in $S^*$ is indicated by the constraints we learn. This property guarantee that $S^*\subseteq \hat{S}.$
            \item Comparing the dimension of $S^*$ and $S$. We find that $d(S^*)=d(\hat{S})=249.$
        \end{compactenum}  
        In this way, we confirm that our framework can mine the underlying Sudoku rules successfully.
        
    \subsection{Minimal Spanning Tree}
    \label{sec:MST}
        As discussed in Sec. \ref{sec:intro}, inference problems in many structured output prediction applications (e.g., dependency parsing) can be modeled as searching the minimal (or maximal) spanning tree (MST). In the following, we verify if the proposed approach can identify the structure of solution is a tree by merely providing pairs of adjacency matrices and the corresponding MST. We encode the MST problem as described in Sec. \ref{sec:method}.

        \begin{table}[t]
            \caption{MST results in exact match (EM) accuracy in Train and Test, the average edge accuracy (Edge) and the ratio of solutions that are feasible (Feasibility). Here, the feasibility means the solution forms a tree.}
                        \small
            \begin{subtable}{\linewidth}
                \centering
                \caption{$10,000$ training samples}
                \begin{tabular}{@{}c|@{\hskip3pt}c@{\hskip3pt}|@{\hskip3pt}c@{\hskip3pt}|@{\hskip3pt}c@{\hskip3pt}|c@{}}
                    \centering
                    Model & Train EM & Test EM & Test Edge & Test Feasibility\\
                    \hline
                    \textsf{NN} & 8.3\% & 6.9\% & 89.3\% & 12.8\%\\
                    \inner & 100\% & 41.6\% & 93.7\% & 100\%\\
                    \mouter & 100\% & 71.1\% & 96.1\% & 71.1\%\\
                    \outereq & 100\% & 72.9\% & 96.1\% & 72.9\%\\
                \end{tabular}
                \label{tab:2-1}
            \end{subtable}
            \begin{subtable}{\linewidth}
                \centering
                        \vspace{5pt}
                \caption{$20,000$ training samples}
                  \begin{tabular}{@{}c|@{\hskip3pt}c@{\hskip3pt}|@{\hskip3pt}c@{\hskip3pt}|@{\hskip3pt}c@{\hskip3pt}|c@{}}
                    Model & Train EM & Test EM & Test Edge & Test Feasibility\\
                    \hline
                    \textsf{NN} & 9.5\% & 10.4\% & 91.1\% & 13.4\%\\
                    \inner & 100\% & 69.2\% & 97.0\% & 100 \%\\
                    \mouter & 100\% & 87.2\% & 98.0\% & 87.2\%\\
                    \outereq & 100\% & 91.8\% & 98.7\% & 91.8\%\\
                \end{tabular}
                \label{tab:2-2}
            \end{subtable}
            \label{tab:mst}
        \end{table}
        
        We generate a dataset with $7$ nodes. The dataset contains $20,000$ training and $500$ test data. Every
        data point consists of an adjacency matrix serialized in a vector $\mathbf{w}$, where $\mathbf{w}_{i,j}$ represents the distance between node $i$ and node $j$, and its corresponding MST. The entry in the adjacency matrix is independently sampled from a uniform distribution in $[-1,1].$ We filtered out the adjacency matrix with identical values to ensure every sample has a unique optimal solution. 
        
        We test the inner polytope (\inner), outer polytope (\mouter) and outer polytope with equality constraint mining (\outereq) methods. 
        We compare our approaches with fully connected feed-forward neural networks (\textsf{NN}) with $1,2$ or 3 layers, which directly learn the association between $\mathbf{w}$ and $\mathbf{y}$ and the hyper-parameters are given in the Appendix. We set the hidden dimension to be $50$. Despite that our methods do not have hyper-parameters, to tune the neural network, we generate another $500$ dev data points.

        Table \ref{tab:mst} shows the results in exact match (i.e., correct MST) and edge accuracies.  The results show that Baseline-NN is unable to learn the tree structure from the given examples. We find that \textsf{NN} can learn reasonably well in each individual edge but is terrible to capture the output is a tree. In particular, in 87.2\% and 86.6\% of cases for 10,000, 20,000 training samples, respectively, the output by \textsf{NN} is not feasible (not a tree). Therefore, its exact match accuracy is low. In comparison, the proposed approaches \mouter and \outereq mostly produce feasible solutions\footnote{Note that for \mouter and \outereq methods, 
        the results of feasibility are the same as test EM because all feasible trees are in the outer polytope. 
        Therefore, if a model outputs a feasible tree, the tree is guaranteed to be optimal.}, resulting in much higher exact match accuracy. 
        Comparing the results for $10,000$ and $20,000$ data points, we find that \outereq is more effective than \textsf{NN} when doubling the training data as it improves 20\% exact match accuracy.

    \subsection{Hierarchical Multi-label Classification}
        \begin{figure}[t]
            \centering
            \includegraphics[width=0.8\linewidth]{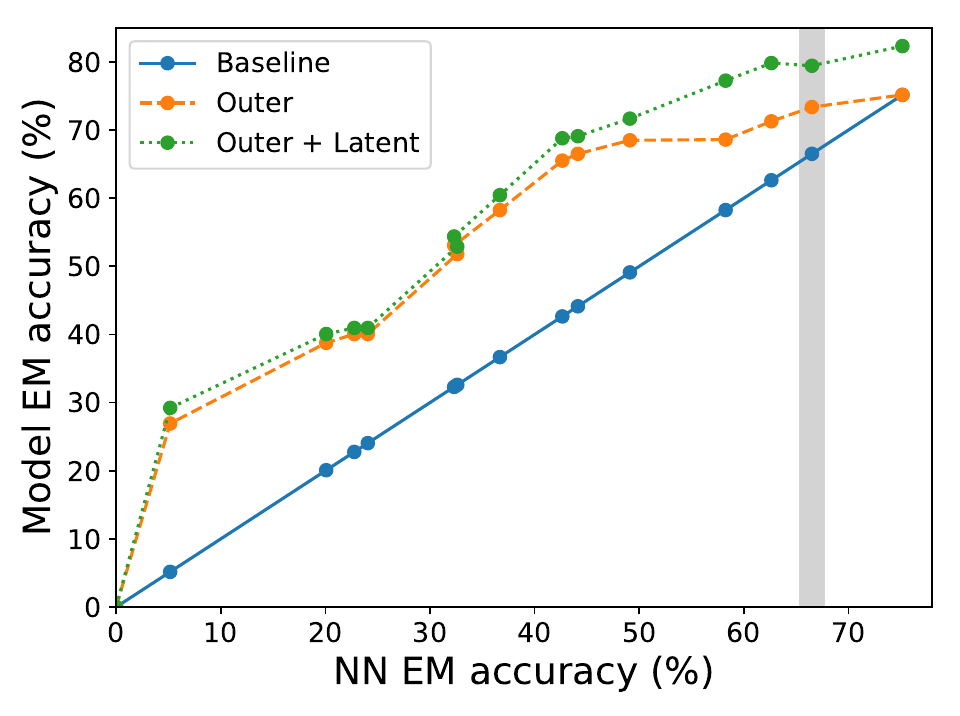}
            \caption{Results on ImCLEF07A in EM accuracy using base models with different performance levels.}
            \label{fig:multi}
        \end{figure}

        \begin{table}[t]
            \centering
            \caption{Detailed results (grey zone in Fig. \ref{fig:multi}) evaluated by exact match accuracy (EM), the ratio of solutions that are feasible (Feasibility), and average accuracy of label assignments (Label Acc.).
            }
            \small
            \begin{tabular}{@{}c|c|c|c@{}}
                Model & Test EM & Test Feasibility & Test Label Acc.\\
                \hline
                \baseline & 62.6\% & 73.8\% & 98.92\% \\
                \inner & 79.8\% & 100\% & 98.98\%\\
                \mouter & 71.3\% & 89.7\% & 98.85\%\\
                \latent & 79.8\% & 100\% & 98.98\%\\
            \end{tabular}
            \label{tab:3-2}
        \end{table}

        Finally, we apply the proposed approaches to a real-world problem and demonstrate its ability to cooperate with machine learning models. We conduct experiments on ImCLEF07A \citep{dimitrovski2011hierarchical}, which contains $10,000$ training samples and $1,006$ test samples. Each sample has $80$ features and a set of labels selected from $96$ classes. There is a hierarchy among the labels and the depth of the hierarchical structure is $4$. A feasible label set forms a path from the root to a leaf node.
        
        The base model is a 3-layer fully connected feed-forward neural network with hidden dimension $80.$ This model outputs a vector $\mathbf{c}$, where each component $c_i \in [0,1]$ is predicted independently to the input instance. For the baseline model, if $c_i>0.5$, then the label $i$ is positive. 
        
        We take the base model as a sub-routine and use it to assign weight $\mathbf{w}$ in Eq. \eqref{eq:ilp}. Specifically, $\mathbf{w}=\mathbf{c}-0.5\times \mathbf{1}.$
        Without constraints, solving the ILP in Eq. \eqref{eq:ilp} is equivalent to make predicton by the baseline model. 
        We evaluate 1) the inner polytope method (\inner), 2) the outer polytope method (\mouter) and 3) the outer polytope method with latent variables (\latent). 
        In \mouter, as $\mathbf{w}$ is generated by a predicted model, we use Eq. \eqref{eq:adaptconstr} to allow noise.
        In \latent, we use the label pairwise latent variables defined in Sec \ref{sec:latent}. To reduce the label spaces, we follow the convention to consider only induce latent variables to label pairs that occur in the training set.

        Different from Sec. \ref{sec:sudoku} and \ref{sec:MST},
        the weight $\mathbf{w}$ is a score vector predicted by the base model. To understand how the constraint mining approaches incorporate with base models with different performance levels, we train multiple versions of base models with different number of layers and training epochs then demonstrate the performance of our approach with these base models. The hyper-parameters for these models are given in the Appendix. The results are shown in Fig. \ref{fig:multi}. 
        
        Results show that \mouter and \latent improve the base models in all cases. Even with a weak base model with only $10\%$ in exact match accuracy, \mouter and \latent are able to learn underlying constraints and improve the performance by more than 20\%. The difference between \mouter and \latent is not apparent when the base model is inaccurate. However, when the base model performance increases, \latent is capable of capturing more fine-grained constraints than \mouter and achieves better performance. 
        When the baseline achieves $0$ loss in training data (the right-most column points), the constraints learned by \mouter can not filter out any point in the space. Therefore, \mouter achieves the same performance as \textit{Baseline}. However, \latent can still mine constraints related to the latent variable and improve the performance.
        
        Table \ref{tab:3-2} highlights the detailed results with one base model.\footnote{We choose the second best baseline model since the best one gets 100\% accuracy in training set, which causes the inequality constraints learned by \mouter method filter out nothing.} \mouter improves \baseline about $12\%$ in exact match accuracy and $16\%$ in feasibility. This demonstrates \mouter can successfully filter out many infeasible solutions and guide the model to find the correct ones. 
        \inner learns exactly the feasible set as all pairs of labels appear in the test set also appear in the training. Similarly, \latent is able to identify all dependencies between labels and achieves high performance. The classes in this task have a tree structure and it has depth $4$ including the root (root is a virtual concept, and it is not a real class). We use $p(x)$ to denote the parent class of class $x$, and $L_i$ to denote the set of classes on layer $i$, $i\in\{1,2,3\}.$  We verify the constraints with the same method in Sec. \ref{sec:sudoku}. The mined equality constraints are the linear transformation of the following constraints:
        \begin{equation}
        \label{eq:hmcconst}
            \begin{split}
                \sum\nolimits_{j\in L_i} y_j &= 1, \forall i\in\{1,2,3\}, \\
                h_{parent(x),x,0,1} &= 0, \forall x:parent(x)\neq root.\\
            \end{split}
        \end{equation}
       
 \section{Analysis and Discussion}       
\subsection{Feasible Set Size Analysis}
\label{sec:analysis}
    We provide a theoretical analysis about the convergence speed of the proposed approaches by estimating the cardinality of the outer polytope and inner polytopes. Our methods in Sec. \ref{sec:method} estimate the feasible set by squeezing the outer polytope and enlarging the inner polytope. We analyze how the sizes of outer polytope and inner polytope change with respect to the number of training samples.
    
    We denote the size of ground truth feasible set $S^*$ as $M$, the size of universal label space is $N$. The weights $\mathbf{w}$ are drawn from the distribution $D_{\mathbf{w}}$. For each point $i$, we use $p_i$ to denote the probability that given a randomly sampled weight $\mathbf{w}\sim D_{\mathbf{w}}$, $\mathbf{y}^{(i)}$ get higher score than all the feasible points. Formally,
        $p_i=\mathbf{\emph{P}}_{\mathbf{w}\sim D_{\mathbf{w}}}\{\mathbf{w}\cdot \mathbf{y}^{(i)}\geq \mathbf{w}\cdot \mathbf{y}^{(j)}, \forall j\in S^*\}.$
    The following lemma bounds the expectation sizes of the outer and inner polygons. Full proof is in the Appendix. 
    
    \paragraph{Lemma 3:} The expectation of the sizes of outer and inner polygon is given by
    \begin{equation}
    \label{eq:expect}
        \begin{split}
            \mathbb{E}[|S_I|]=M-\sum\nolimits_{i\in S^*}\left(1-p_i\right)^k;
            \\
            \mathbb{E}[|S_O|]=M+\sum\nolimits_{j\notin S^*}(1-p_j)^k.
        \end{split}
    \end{equation}
    
    \paragraph{Proof sketch:} We first consider the outer polygon. The point $j$ out of the feasible set appears in the outer polygon if and only if it is not filtered out by any constraints, which is 
    $(1-p_j)^k.$
    
    We then consider the inner polygon. The point $i$ appears in the inner polygon if and only if at least one training sample takes it as the optimal solution, which is $1-(1-p_i)^k.$
    
    
    
    
    
    \paragraph{Case study: MST} We take MST discussed in Sec.\ref{sec:MST} as an example. According to Matrix-Tree Theorem \citep{chaiken1978matrix}, we know that the number of spanning trees is $M=16,807.$ The universal set size before mining equality constraints is $N_O=2^{21}=2,097,152$. However, with the equality constraints mining method, we can identify the constraint $\sum_{i=1}^{21} y_i=6,$ (i.e., number of edges is $6$). With this constraint the size of the space is reduced to $N=\binom{21}{6}=54,264.$
    
\begin{figure}[t]
        \begin{subfigure}{0.48\linewidth}
            \centering
            \includegraphics[width=\linewidth]{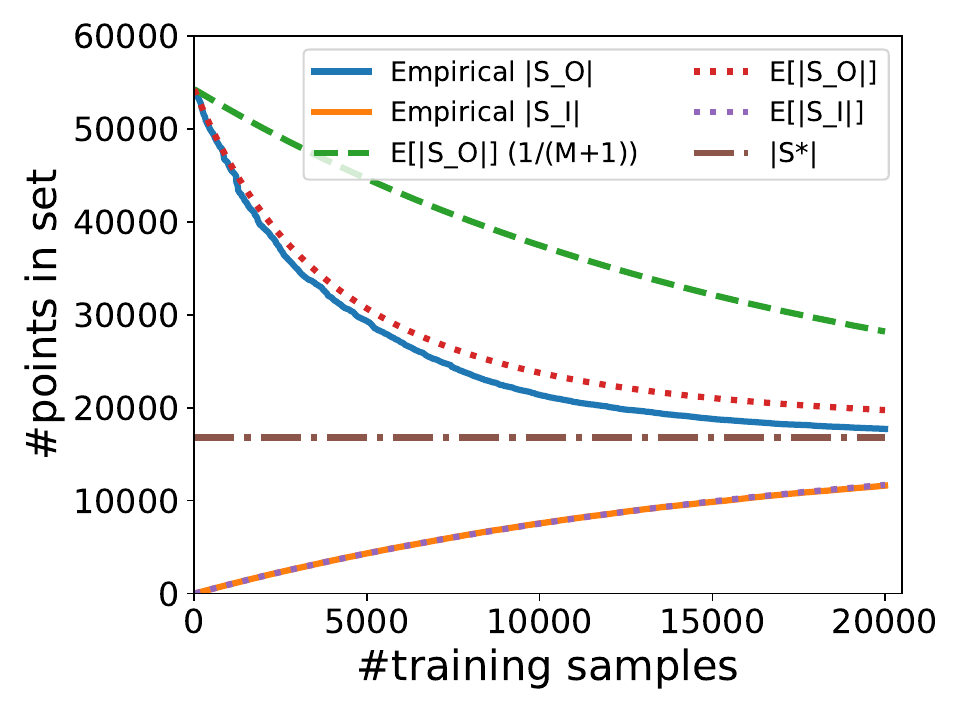}
            \caption{Set size}
            \label{fig1}
        \end{subfigure}
        \hspace{0.01in}
        \begin{subfigure}{0.48\linewidth}
            \centering
            \includegraphics[width=\linewidth]{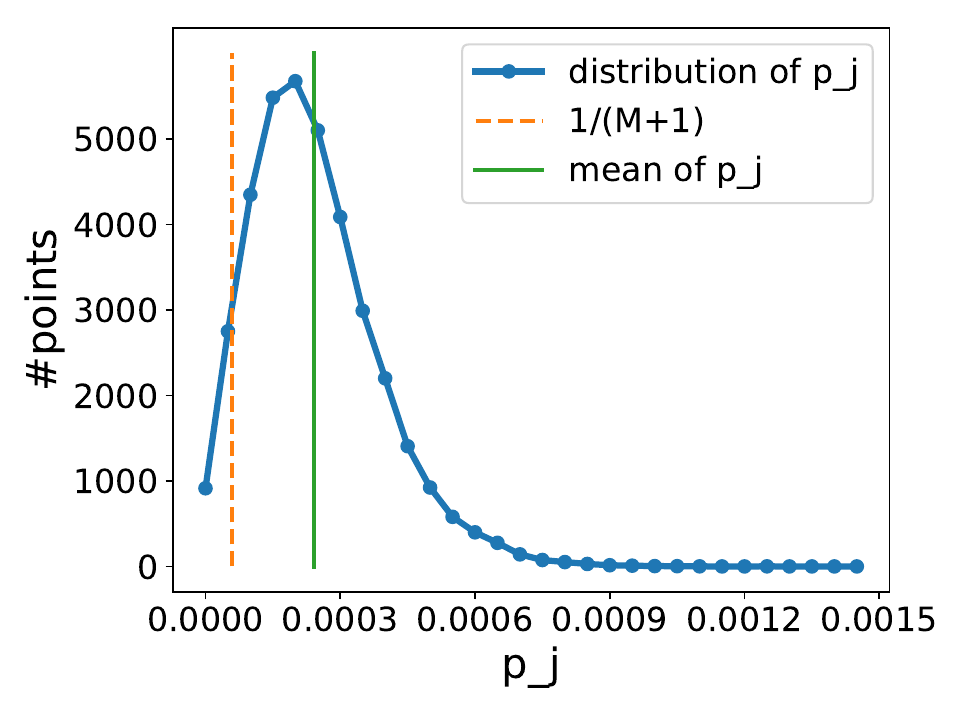}
            \caption{The dist. of $p_j,j\notin S^*$}
            \label{fig2}
        \end{subfigure}
        \caption{On the left Fig. \ref{fig1} shows the empirical and expectation sizes of outer and inner polytope, comparing with ground truth. $E[|S_O|](1/(M+1))$ is the expectation of outer polytope size estimated by $p_j=\frac{1}{M+1}$ while $E[|S_O|]$ is estimated with empirical $p_j$. Note that the empirical inner polytope and its theoretical curve almost coincide. On the right Fig. \ref{fig2} shows the distribution of $p_j$ estimated from $20,000$ training data comparing with $1/(M+1).$}
    \end{figure}

    $p_i$ in Eq. \eqref{eq:expect} is difficult to estimate directly; therefore, we approximate it by $p_i=\frac{1}{M}$ for $i\in S^*$ and $p_j=\frac{1}{M+1}$ for $j\notin S^*$. The approximation is exact if the following assumption holds  (see details and proof in Appendix).
    \emph{Data symmetric}: for $k$ different points $\mathbf{y}^{(1)},\mathbf{y}^{(2)},\dots,\mathbf{y}^{(k)}$, $\mathbf{\emph{P}}_{\mathbf{w}\sim D_{\mathbf{w}}}\{\mathbf{w}\cdot \mathbf{y}^{(1)}\geq \mathbf{w}\cdot \mathbf{y}^{(i)}, i\in [k]\}=\frac{1}{k}.$
    MST only satisfies the part of the assumption, therefore, for $i\in S^*$, the approximation $p_i=\frac{1}{M}$ is close, and there is a gap between empirical $p_j (j\notin S^*)$ compared with $\frac{1}{M+1}$ (see Fig. \ref{fig2}), this causes the gap between the empirical result and the estimated expectation about the outer polytope. Fig. \ref{fig1} shows the empirical sizes of outer and inner polytope $|S_O|,|S_I|$, with their theoretical expectation in Eq. \eqref{eq:expect} and the ground truth $|S^*|$. The expectation of \inner perfectly fits the empirical results and the two curves are almost coincide. Both of the \mouter and \inner methods eventually converge to the ground truth, and the \mouter method is closer.

\subsection{Discussion about Running Time}
\label{sec:time}
    There are two main factors affecting the running time: the number of data samples, the number of constraints $K$, and the size of the output variables $D$. Our approach contains two steps: identifying the feasible set (training) and solving ILPs (inference). In training, as shown in Sec \ref{sec:method}, our approach is linear in $K$ since we only needs to pass all samples once, and no worse than quadratic in $D$. In inference, we solve ILP which is generally NP-hard, and the main factor in complexity is $D$. However, for most structured prediction tasks, $D$ is small. In our experiments, $D$ is $21$ and $729$ and $96$ in MST, Sudoku and HMC, respectively.
    
    Table \ref{tab:rt} shows the training and test running time of our approaches compared to neural network models. In training, our approach is more efficient than the baseline neural network in the Sudoku and MST experiments. For the HMC experiment, the training time of our approach includes updating the model parameters of the underlying neural networks. Therefore, the training time is longer compared to the Sudoku and MST cases.

    \begin{table}[t]
        \centering
        \caption{Running time in seconds for experiments. The training time is computed by averaging in 3 runs, while inference time is computed by averaging in 100 samples.}
        \small
        \begin{tabular}{c|c|c|c}
            Experiment & Model & Training & Inference\\
            \hline
            \multirow{3}{*}{MST} & \baseline NN & 190 & 5e-4 \\
            & \mouter & 24.1 & 7.9 \\
            & \outereq & 26.5 & 3.8 \\
            \hline
            \multirow{2}{*}{Sudoku} & ConvNet & 636.4 & 5e-4 \\
             & \outereq & 5.2 & 1.2 \\
            \hline
            \multirow{3}{*}{HMC} & \baseline NN & 44.9 & 4e-4\\
            & \mouter & 560.3 & 1.1 \\
            & \latent & 599.3 & 7.1 \\
            \hline
        \end{tabular}
        \label{tab:rt}
    \end{table}

    
    For the inference time, we report the average time on solving one test sample. Despite ILP is NP-hard, a commercial solver (e.g., Gurobi) is capable of solving the problem within a reasonable time. Therefore, without carefully engineering to optimize the running time, the ILP solver can produce solutions within a few seconds. 
    
\begin{figure}[t]
        \centering
        \includegraphics[width=0.8\linewidth]{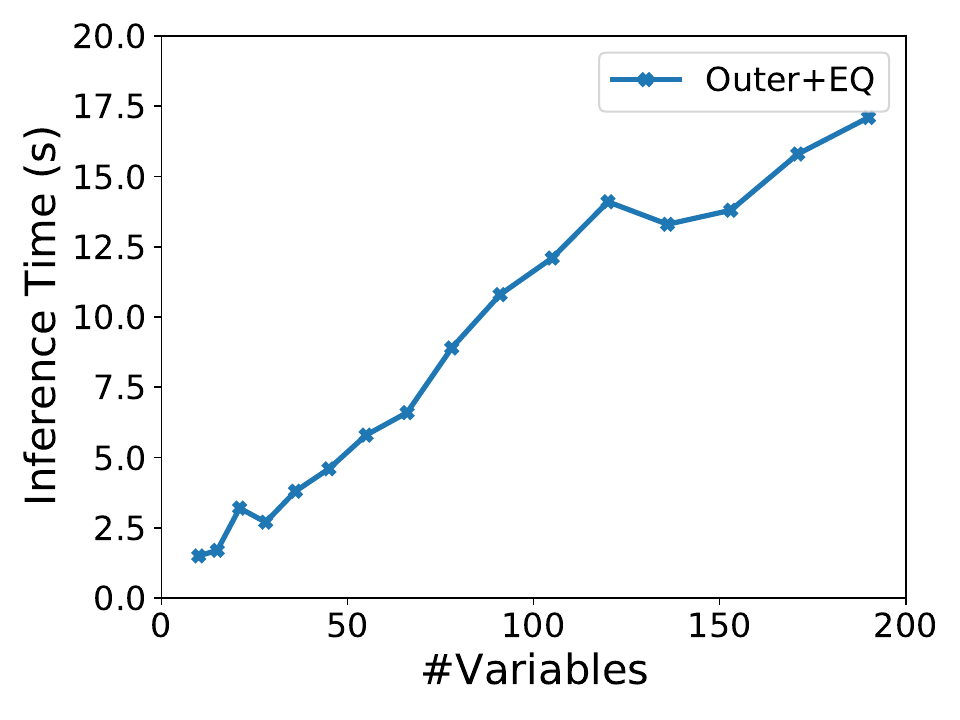}
        \caption{The inference time per instance on MST experiments with different number of variables. Here x-axis is the number of variables $D$, which equals to $N(N-1)/2$. $N$ is the number of nodes in the graph. y-axis is the inference time, computed by averaging 100 samples.}
        \label{fig:size-time}
    \end{figure}

    To empirically understand the scalability of our approach in the inference, we test the inference time in the MST experiment with larger graph. Here we fix the number of constraints (number of training sampels) as $20,000$. The results are shown in Fig. \ref{fig:size-time}. We find that despite some extreme cases, the inference time grows generally linearly in terms of the number of variables empirically.
    
    Note that although the constraints are mined using the ILP framework, it does not mean that the inference has to be solved by an ILP solver. Once the constraints are identified, one can design a specific constraint solver to speed up the inference. Besides, the ILP inference can be accelerated by amortizing the computations when solving a batch test instances~\citep{srikumar2012amortizing, chang2015structural} or by applying approximate inference algorithms for solving ILP, e.g., LP relaxation methods \citep{kulesza2007structured, martins2015ad3}.

    \begin{figure}[t]
        \centering
        \begin{subfigure}{0.48\linewidth}
            \includegraphics[width=\linewidth]{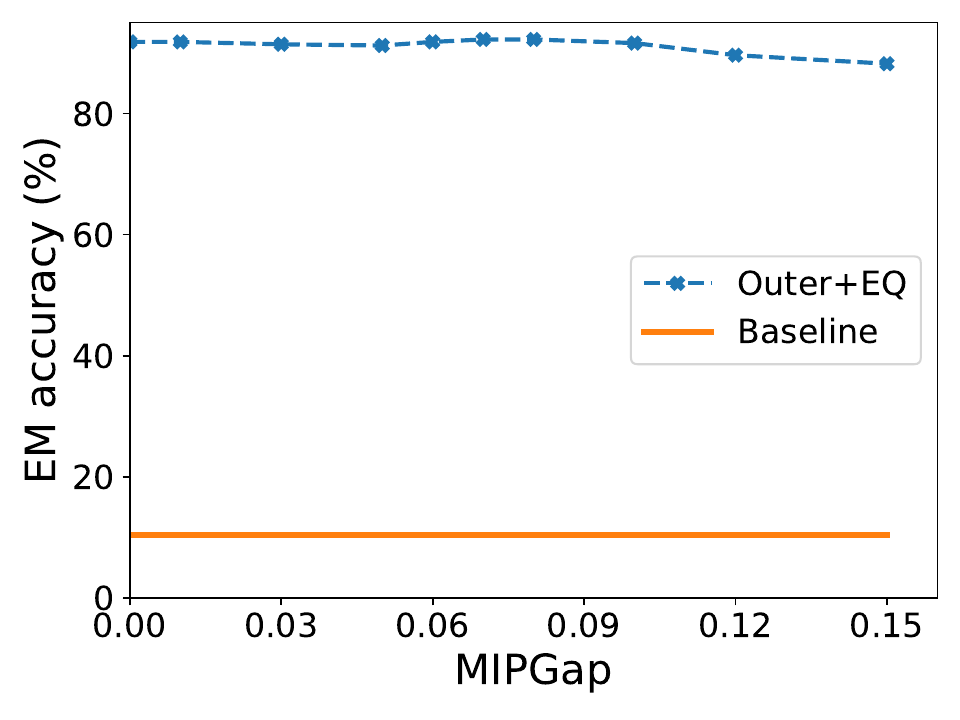}
            \caption{EM accuracy}
            \label{fig:time-1}
        \end{subfigure}
        \begin{subfigure}{0.49\linewidth}
            \includegraphics[width=\linewidth]{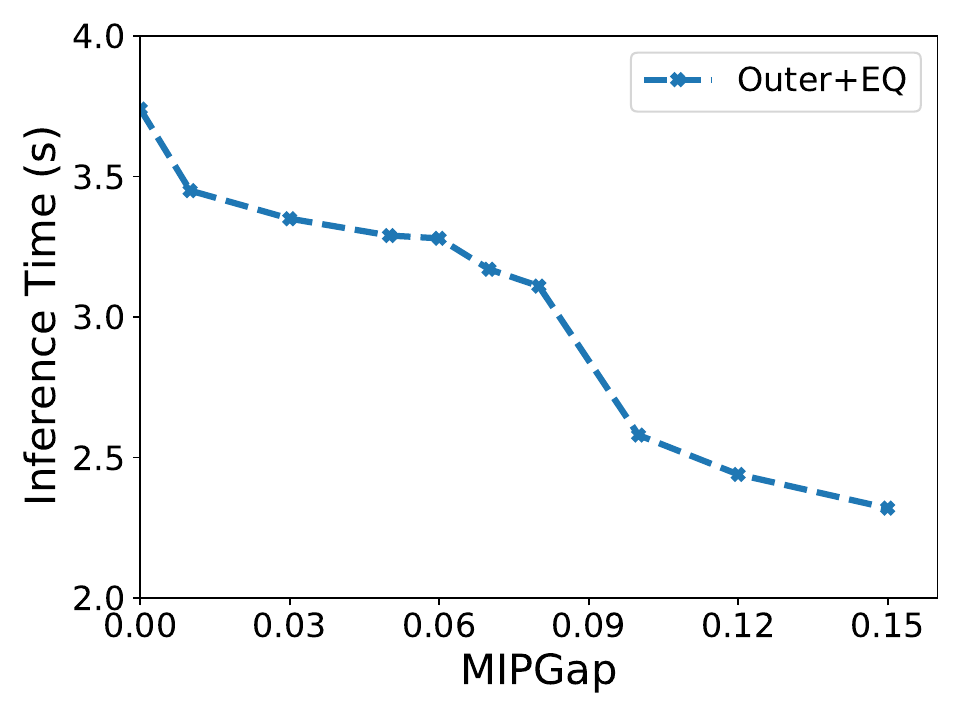}
            \caption{Inference time}
            \label{fig:time-2}
        \end{subfigure}
        \caption{The trade-off between inference time and model accuracy when an approximate inference solver is used. Results are on MST. The x-axis is the allowed maximum relative gap between the returned solution and the optimum solution. On the left figure, the performance of our approach drops when the MIPGap is large, but our approach  still significantly outperforms the neural network baseline (10.4\%). The right figure shows that the inference time significantly reduces when MIPGap gets large.}
        \label{fig:time}
    \end{figure}
    
    To demonstrate how the performance of our approach is affected by the approximate ILP solver. 
    We show a trade-off curve in MST 20,000 training experiments in Fig. \ref{fig:time} by solving inference using Gurobi with different MIPGap, a parameter of the Gurobi solver controlling the quality of solutions. Specifically, MIPGap specifies the maximum gap of the objective function values between the returned solution and the optimum solution. We vary MIPGap from 0 (exact solutions are returned) to 0.15. 
    The experimental results demonstrate that the inference can be accelerated using an approximate inference solver with a trade-off of moderate performance loss.


\section{Conclusion}
    We propose an integer linear programming framework for mining constraints from data. The framework is general and is able to identify underlying constraints in structured prediction problems. 
    Experiments on synthetic problems and hierarchical classification show that the framework is capable of mining complex constraints over a label space, and it can cooperate with neural models. As the first paper to formulate the constraints mining as ILP, we focus on building the foundation for this potential area and understanding the properties of the proposed approach.
    
\section{Acknowledgement}
    This work was supported by National Science Foundation Grant IIS 1927554 and a Facebook Research Award. We appreciate Cheng Ma and members of the UCLA-NLP lab for their inputs and feedback during this project. We also thank the anonymous reviewers their valuable comments.

\bibliography{icml_2021}
\bibliographystyle{icml_2021}

\clearpage
\appendix
\section{Mined Equality Constraints in Sudoku Experiments}
    \label{app:sudoku}
    The mined equality constraints are the linear transformation of the following constraints:
    \begin{equation}
    \label{eq:sudokuconst}
        \begin{split}
            \sum_{i=1}^9 y_{ijk} &= 1,\ \forall j,k,\\ 
            \sum_{j=1}^9 y_{ijk} &= 1,\ \forall i,k,\\ 
            \sum_{k=1}^9 y_{ijk} &= 1,\ \forall i,j,\\ 
            \sum_{i=1}^3 \sum_{j=1}^3 y_{(x+i)(y+j)k} &= 1, \forall x,y\in \{0,3,6\} \forall k.
        \end{split}
    \end{equation}
    These are the rules of Sudoku described by linear constraints. We denote the linear space defined by these constraints as $S^*$, and the linear space given by our mined constraints as $\hat{S}$. We verify $S^*=S$ by
    \begin{enumerate}
        \item For each constraint in Eq. \eqref{eq:sudokuconst}, we verify it is indicated in $\hat{S}$. This property guarantee that $S^*\subseteq \hat{S}.$
        \item Comparing the dimension of $S^*$ and $S$. We find that $d(S^*)=d(\hat{S})=249.$
    \end{enumerate}  

\section{Mined Equality Constraints in Hierarchical Multiclass Classification Experiments}
    \label{app:hmc}
    The classes in this task have a tree structure and it has depth $4$ including the root (root is a virtual concept that it is not a real class). We use $p(x)$ to denote the parent class of class $x$, and $L_i$ to denote the set of classes on layer $i$, $i\in\{1,2,3\}.$ The mined equality constraints are the linear transformation of the following constraints:
    \begin{equation}
    \label{eq:hmcconst}
        \begin{split}
            \sum_{j\in L_i}, y_j &= 1, \forall i\in\{1,2,3\}, \\
            h_{parent(x),x,0,1} &= 0, \forall x:parent(x)\neq root.\\
        \end{split}
    \end{equation}
    We use the same method in Appendix. \ref{app:sudoku} to verify it.

\section{The Expectation of the Size of Outer and Inner Polytopes}
\label{app:size}
    We define 
    $$p_i=\mathbf{\emph{P}}_{\mathbf{w}\sim D_{\mathbf{w}}}\{\mathbf{w}\cdot \mathbf{y}^{(i)}\geq \mathbf{w}\cdot \mathbf{y}^{(j)}, \forall j\in S^*\}.$$
    When $i\in S^*$, the condition in $p_i$ means the under the given weight $\mathbf{w}$, $i$ is the optimal point. So $p_i$ is the probability of $i$ is the ground truth label for a random weight $\mathbf{w}$ sampled from $D_\mathbf{w}.$
    
    We then consider $E[S_I].$ The inner polytope is a convex hull of seen feasible points. For each feasible point $i\in S^*$, the probability that we have seen it after $k$ samples is $1-(1-p_i)^k.$ Thus, after $k$ training samples, the expectation size of the inner polytope is given by
    $$E[S_I]=\sum_{i\in S^*} (1-(1-p_i)^k)=M-\sum_{i\in S^*}(1-p_i)^k,$$
    where $M=|S^*|.$
    
    When $i\notin S^*$, the condition in $p_i$ means the given weight $\mathbf{w}$, $i$ is better than all the feasible points. Given $\mathbf{w}$ in training, we find the label is worse than $i$, which indicate $i$ is infeasible and it will be filtered out. So $p_i$ is the probability that infeasible point $i$ is not filtered out in training for a random weight $\mathbf{w}$ sampled from $D_\mathbf{w}.$
    
    We then consider $E[S_O].$ The outer polytope is initialized as the whole space and filters out infeasible points in training. For each infeasible point $i\notin S^*$, the probability that it is not filtered out after $k$ samples is $(1-p_i)^k.$ We also know that all the feasible points will not be filtered out. Thus, after $k$ training samples, the expectation size of the outer polytope is given by
    $$E[S_O]=M+\sum_{i\notin S^*} (1-p_i)^k,$$
    where $N$ is the size of the universal set.
    
\section{Expectation Approximation under Assumptions}
\label{app:app}
    The data symmetric assumption is: for $k$ different points $\mathbf{y}^{(1)},\mathbf{y}^{(2)},\dots,\mathbf{y}^{(k)}$, $$\mathbf{\emph{P}}_{\mathbf{w}\sim D_{\mathbf{w}}}\{\mathbf{w}\cdot \mathbf{y}^{(1)}\geq \mathbf{w}\cdot \mathbf{y}^{(i)}, i\in [k]\}=\frac{1}{k}.$$
    With this assumption, consider $p_i=\mathbf{\emph{P}}_{\mathbf{w}\sim D_{\mathbf{w}}}\{\mathbf{w}\cdot \mathbf{y}^{(i)}\geq \mathbf{w}\cdot \mathbf{y}^{(j)}, \forall j\in S^*\}.$ When $i\in S^*$, there are $M$ points are taken into consideration. With the assumption, we can get $p_i=\frac{1}{M},i\in S^*.$ When $i\notin S^*$, there are $M+1$ points are taken into consideration. With the assumption, we can get $p_i=\frac{1}{M+1},i\notin S^*.$ 

\section{Configurations for the Reproducibility}
\label{app:exp}
    \paragraph{Data}
    All the date and code can be found in \url{https://github.com/MtSomeThree/ILPLearning}.
    
    \paragraph{Sudoku Experiments} 
    In the Sudoku experiments, we use the baseline following the settings in \textsf{SATNet}\citep{wang2019satnet} \footnote{The baseline models can be found in \url{https://github.com/locuslab/SATNet}.}.
    
    \paragraph{MST Experiments} 
    In the MST experiments, we use the 3-layer feedforward neural network as the baseline model with ReLU activation. The hidden dimension is set to be $50$. The input and output dimension is $21.$ We use the sigmoid function to regularize the output in $(-1,1).$ We train the model for $300$ epochs and we use the Adam optimizer\citep{kingma2014adam} to optimize the model. The learning rate is set to be $0.001.$  
    
    \paragraph{Hierarchical Multi-label Classification Experiments}
    In the HMC experiments, we use multiple base models. We enumerate the number of layers in $\{1,2,3\},$ the number of training epochs in $\{1,5,50,300\}.$, and the learning rate in $\{0.001, 0.0003, 0.0001\}.$ The hidden dimension is set to be $100.$ The input dimension is $80$ and the output dimension is $96.$ In hidden layer we use ReLU as the activation and in output we use sigmoid function to regularize the output.

\end{document}